\documentclass{article}

\usepackage{graphicx}%
\usepackage{multirow}%
\usepackage{amsmath,amssymb,amsfonts}%
\usepackage{amsthm}%
\usepackage{mathrsfs}%
\usepackage[title]{appendix}%
\usepackage{xcolor}%
\usepackage{textcomp}%
\usepackage{manyfoot}%
\usepackage{booktabs}
\usepackage{algorithm}%
\usepackage{algorithmicx}%
\usepackage{algpseudocode}%
\usepackage{listings}%
\usepackage{hyperref}
\usepackage[T1]{fontenc}
\usepackage{natbib} 
\usepackage{caption}   
\usepackage{threeparttable}  

\title{An Exploratory Analysis on the Explanatory Potential of Embedding-Based Measures of Semantic Transparency for Malay Word Recognition}
\author{
    Mirrah Maziyah Mohamed\thanks{Corresponding author. ORCID: \href{https://orcid.org/0000-0002-3164-3805}{0000-0002-3164-3805}} \\ 
    University of Tübingen, Tübingen, Germany\\ 
    \texttt{maziyah.mohamed@uni-tuebingen.de}  
    \and
    R. Harald Baayen\thanks{ORCID: \href{https://orcid.org/0000-0003-3178-3944}{0000-0003-3178-3944}} \\ 
    University of Tübingen, Tübingen, Germany\\ 
    \texttt{harald.baayen@uni-tuebingen.de} 
}

\date{}

\begin{document}

\maketitle 

\abstract{Studies of morphological processing have shown that semantic transparency is crucial for word recognition. Its computational operationalization is still under discussion. Our primary objectives are to explore embedding-based measures of semantic transparency, and assess their impact on reading. First, we explored the geometry of complex words in semantic space. To do so, we conducted a t-distributed Stochastic Neighbor Embedding clustering analysis on 4,226 Malay prefixed words. Several clusters were observed for complex words varied by their prefix class. Then, we derived five simple measures, and investigated whether they were significant predictors of lexical decision latencies. Two sets of Linear Discriminant Analyses were run in which the prefix of a word is predicted from either word embeddings or shift vectors (i.e., a vector subtraction of the base word from the derived word). The accuracy with which the model predicts the prefix of a word indicates the degree of transparency of the prefix. Three further measures were obtained by comparing embeddings between each word and all other words containing the same prefix (i.e., centroid), between each word and the shift from their base word, and between each word and the predicted word of the “Functional Representations of Affixes in Compositional Semantic Space” model. In a series of Generalized Additive Mixed Models, all measures predicted decision latencies after accounting for word frequency, word length, and morphological family size. The model that included the correlation between each word and their centroid as a predictor provided the best fit to the data.\\

{\bf Keywords}: semantic transparency, embeddings, morphology, lexical decision, Malay
}

\section{Declaration}\label{sec1}

The authors have no competing interests to declare that are relevant to the content of this article. This research was funded in part by the European Research Council, grant \#101054902 (SUBLIMINAL) awarded to Harald Baayen. Data are available \url{https://osf.io/dhyzb/?view_only=e05e71b31cb54daf94a55f46f9cc82da}

\clearpage

\section{Introduction}\label{sec2}

A remarkable phenomenon in language processing in skilled readers is the ability to rapidly decode and extract meaning from written words. A growing body of research on semantic transparency addresses the ease with which a word's meaning is understood, with greater degrees of transparency associated with easier word recognition (e.g., \citealp {chee2022there}; \citealp{diependaele2009semantic}; \citealp{feldman2002semantic}; \citealp{jared2017effect}; \citealp{libben2003compound}; \citealp{marelli2015affixation}). Semantic transparency is typically defined in terms of compositionality, that is, the extent to which the meaning of a complex word can be predicted from the meaning of each of its constituents. From a decompositional perspective of morphological processing \citep{taft1975lexical}, \textit{adapatable} is transparent because the morphemes \textit{adapt} \texttt{+} \textit{-able} describes something or someone that possesses the ability to \textit{adapt}, whereas \textit{moonshine} is fairly opaque as it refers to a type of liquor, rather than following straightfowardly from the meanings of \textit{moon} and \textit{shine}. Word and paradigm, or realizational morphology, offers an alternative explanation in which the word itself represents the most basic unit and that the relationship of words is governed by rules of analogy (e.g., \citealp{hockett1954two}; \citealp{blevins2016word}). In this case, \textit{moonshine} is regarded as semantically opaque because its features are neither related to those of the whole words \textit{moon} nor \textit{shine}. A point of departure between the two main approaches is whether or not there is an explicit representation of morphemes. In more recent distributed accounts of morphological processing (e.g., \citealp{baayen2011amorphous}; \citealp{baayen2019discriminative}; \citealp{gonnerman2007graded}; \citealp{plaut2000non}; \citealp{rueckl2011computational}), typically implemented in the form of connectionist models, morphemes are not explicitly represented. Instead, the representations of a word's form and meaning are shaped by its distributional properties such as the statistical co-occurrences between spelling and meaning.

\setlength{\parskip}{16pt}

There is no clear consensus yet in the operational definition of semantic transparency because word meaning can be studied in various ways (for details on experimental discrepancies, see \citealp{auch2020conceptualizing}). In many of such studies, researchers have relied on human participant ratings, a method that is relatively labor intensive. Here, we explore semantic transparency using multidimensional word embeddings. \citet{westbury2024principal} observed that the initial idea of using high-dimensional matrices to represent word meaning traces back to \citet{osgood1957measurement}, despite the lack of computing power at the time. To represent meaning numerically in a high-dimensional space is, therefore, not entirely a new concept, but rather, a technique that has been refined over time (e.g., \citealp{landauer1997solution}; \citealp{lund1996hyperspace}). Moreover, evidence from \citet{bruni2014multimodal} suggests that the semantic relatedness of words represented by embeddings and human ratings are comparable. The present study capitalizes on recent computational advances to explore the use of high-dimensional word embeddings that may capture word meaning more comprehensively and possibly offer a greater ecological validity compared to small-scale participant ratings. 

\setlength{\parskip}{16pt}
A primary goal for the present study is to further facilitate studies of Malay word recognition. The Malay language, or \textit{Bahasa Melayu}, is a relatively understudied Austronesian language spoken in many regions of Southeast Asia such as Singapore, Malaysia, Brunei, and Indonesia. Malay is rich in derivational morphology with minimal inflection. Derivational affixes are typically used to form words that are related in meanings (e.g., \textit{'baik' good, 'kebaikan' a good action/well-being}). To accomplish the goal of the present study, our first objective is to further augment the Malay Lexicon Project 3, a morphological database, to include a variety of semantic properties for words in the database. At present, the Malay Lexicon Project 3 has estimates of orthographic-semantic consistency calculated for a large subset of simple words. In this study, we calculated several measures that estimate the degree of semantic transparency for a large subset of complex words which will be added to the database. The secondary objective is to evaluate whether, and how well, each measure predicts response times.

\section{Present Study}\label{sec3}

First, we explore the geometry of semantic transparency of complex words in a high-dimensional semantic space and describe our calculations of several measures of semantic transparency that use word embeddings. Then, we evaluated each measure by determining whether they predict lexical decision latencies in a series of Generalized Additive Mixed Models (GAMM).

\subsection{Semantic Geometry of Derived Words}\label{subsec3}

A technique that has been gaining traction and typically used in areas of machine learning to visualize high-dimensional data is the t-distributed stochastic neighbor embedding clustering analysis (t-SNE; \citealp{van2008visualizing}). A t-SNE is an unsupervised nonlinear dimensionality reduction technique. A key insight from Distributional Semantics is that semantically related words appear in similar contexts. As such, semantically related words have similar embeddings that appear closer in the t-SNE space than other words. t-SNE has been used successfully to explore productivity and semantic transparency of German \citep{stupak2022inquiry} and Mandarin \citep{shen2022adjective}. Of particular relevance, complex words in German show clustering by derivational suffix, but not by particles. Similarly, in Mandarin, clusters were detected for suffixes. Importantly, clusters in a t-SNE are driven by information that is most saliently encoded in the embeddings. 

In this study, we adopted this technique to explore the morphological structure of Malay prefixed words. Recent computational \citep{denistia2022morphology} and corpus-based work \citep{denistia2022exploring} on Indonesian prefixation, an Austronesian language closely related to Malay, have shown that embeddings are informative in discriminating the semantics of prefixes \textit{pe-} and \textit{peN-}. Pre-trained 300-dimensional word embeddings were first extracted from FastText \citep{bojanowski2017enriching} for all words in the MLP database for which there are embeddings. Of those, 4,226 words containing at least one of 10 prefixes were analyzed using the {\it Rtsne} package \citep{krijthe2015rtsne} in R. The resulting output are the spatial coordinates for each word in two dimensions as depicted in Figure 1. Considerable clustering is revealed for a variety of words colour-coded by their derivational prefix, except for words containing {\it peri-} (n=3), {\it pra-} (n=4), {\it pe-} (n=35), and {\it se-} (n=31) for which there are too few of such words in our dataset for meaningful clustering, if any, to occur. \textit{MeN-} words (cyan) appear largely in the middle and represent the majority of the data. \textit{BeR-} words (yellow) cluster in two groups in the bottom right. \textit{TeR-} words (purple) mostly appear on the outskirts of meN- words in the middle. \textit{PeN-} words (red) appear on the edges forming an outer ring. \textit{PeR-} words (teal) are sparsely scattered only on the left. \textit{Ke-} words (navy blue) cluster in two groups in the bottom left. Importantly, across a variety of prefixes, we observe a strong form-meaning correspondence such that words containing the same prefix appear closer in semantic space to each other than words that contain different prefixes.

\begin{figure}[h] 
    \centering
    \includegraphics[width=0.8\textwidth]{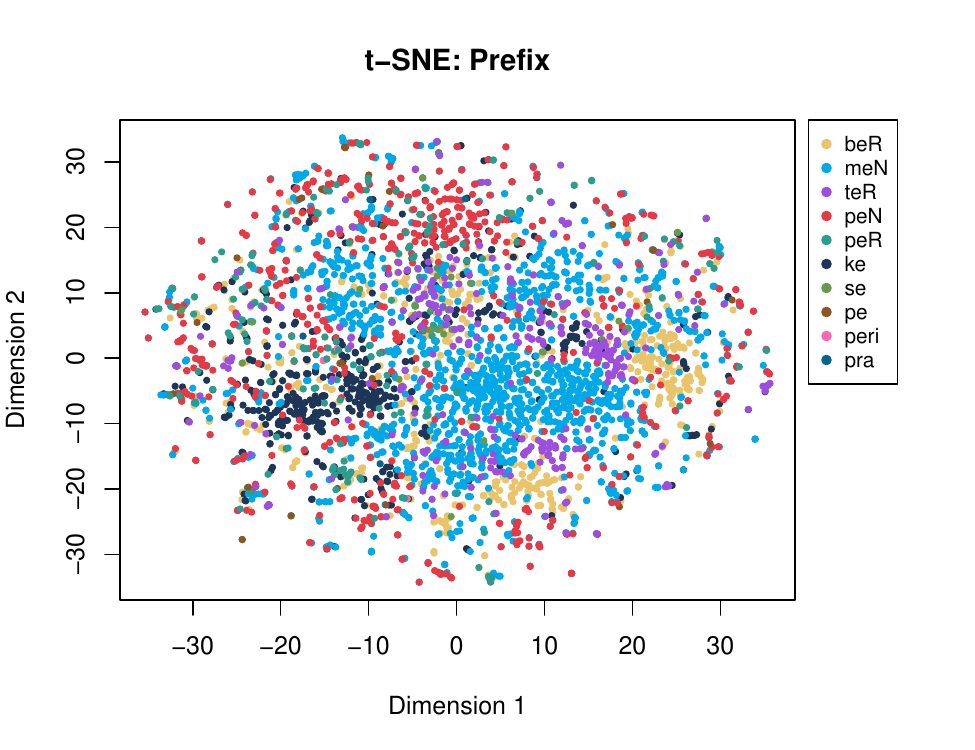}
    \caption{{\it Note}. Each coloured dot represents a derived word and each colour corresponds to a particular prefix a word contains. Words that have similar embeddings appear closer to each other in the t-SNE space than others.}
    \label{fig1}
\end{figure}

\subsection{Embedding-based Measures}\label{subsec4}

\subsubsection{Linear Discriminant Analysis}\label{subsubsec4}

Following the t-SNE analysis, we computed two measures of semantic transparency of prefixes by conducting Linear Discriminant Analyses (LDA). LDA is an approach used in supervised machine learning to solve classification problems. Two sets of LDA models were run, each with a different input, and examined whether each word is linearly separable in the embedding space by their prefix. One dataset contained the vectors for each derived word. The second dataset contained shift vectors for each word, that is, the displacement in semantic space of the derived word from its base by subtracting their vectors. Importantly, in both cases, the accuracy with which the model successfully predicts a word’s prefix is an index for the degree of correspondence between the form and meaning of a prefix. The predictions of the LDA model that are derived from a leave-one-out cross-validation approach are presented in Tables 1 and 2. Results from the LDA are in tandem with those of the t-SNE, that is, the LDA produced mostly correct classifications for words that contained prefixes that cluster successfully in the t-SNE. As an example, of 754 words that contained the prefix \textit{beR-}, the LDA model accurately classified words containing the prefix \textit{beR-} 713 times, yielding an accuracy of .946 for the prefix \textit{beR-}. A cautious approach to the interpretation of the LDA results is to compare the proportion of correct classifications for each prefix against a baseline accuracy. The baseline accuracy in this case is .40, and can be calculated by taking the number of words that contain the prefix that occurs the most (i.e., \textit{meN-}) divided by the total number of words. Overall, both LDA models predicted class membership accurately (.93 using derived word vectors, and .88 using shift vectors), providing evidence for the effectiveness of word embeddings and their shift vectors in discriminating words of different prefixes. 

\begin{table}[h]
    \caption{Predictions of Class Membership using Embeddings}\label{tab1}
    \centering
    \resizebox{\textwidth}{!}{ 
    \begin{tabular}{rrrrrrrrrrrrr}
    \toprule
    & beR & ke & meN & pe & peN & peR & peri & pra & se & teR & Total & Accuracy \\
    \midrule
    beR & {\bf 713} & 4 & 21 & 0 & 1 & 1 & 2 & 0 & 0 & 9 & 754 & .946 \\
    ke & 5 & {\bf 456} & 2 & 0 & 16 & 25 & 0 & 0 & 3 & 2 & 509 & .896 \\
    meN & 15 & 5 & {\bf 1649} & 0 & 0 & 2 & 0 & 0 & 4 & 5 & 1680 & .982 \\
    pe & 0 & 7 & 0 & {\bf 14} & 10 & 4 & 0 & 0 & 0 & 0 & 35 & .400 \\
    peN & 1 & 20 & 0 & 7 & {\bf 618} & 31 & 0 & 0 & 1 & 0 & 678 & .912 \\
    peR & 2 & 22 & 1 & 2 & 15 & {\bf 155} & 0 & 1 & 1 & 1 & 200 & .775 \\
    peri & 0 & 2 & 0 & 0 & 0 & 1 & {\bf 0} & 0 & 0 & 0 & 3 & .000 \\
    pra & 0 & 0 & 0 & 0 & 1 & 1 & 0 & {\bf 2} & 0 & 0 & 4 & .500 \\
    se & 5 & 1 & 3 & 0 & 0 & 0 & 0 & 0 & {\bf 20} & 2 & 31 & .645 \\
    teR & 14 & 1 & 3 & 0 & 1 & 1 & 0 & 0 & 1 & {\bf 311} & 332 & .937 \\
    \bottomrule
    \end{tabular}
}
    \begin{tablenotes}
        \small
        \item {\it Note.}The prefix in each row represents the prefix of a word and the prefix in each column represents the predicted prefix of a word. The bolded values represent the number of correct classifications of the LDA model. The overall accuracy is .93
    \end{tablenotes}
\end{table}

\begin{table}[H]
    \caption{Predictions of Class Membership using Shift Vectors}\label{tab2}
    \centering
    \resizebox{\textwidth}{!}{ 
    \begin{tabular}{rrrrrrrrrrrrr}
    \toprule
    & beR & ke & meN & pe & peN & peR & peri & pra & se & teR & Total & Accuracy \\
    \midrule
    beR & {\bf 595} & 6 & 23 & 0 & 4 & 4 & 0 & 0 & 3 & 11 & 646 & .921 \\
    ke & 15 & {\bf 396} & 4 & 3 & 22 & 26 & 0 & 0 & 1 & 4 & 471 & .841 \\
    meN & 32 & 4 & {\bf 1290} & 0 & 4 & 2 & 0 & 0 & 5 & 6 & 1343 & .961 \\
    pe & 2 & 6 & 0 & {\bf 11} & 10 & 1 & 0 & 0 & 0 & 0 & 30 & .367 \\
    peN & 5 & 24 & 0 & 5 & {\bf 503} & 33 & 0 & 0 & 1 & 1 & 572 & .879 \\
    peR & 8 & 29 & 1 & 2 & 30 & {\bf 100} & 0 & 0 & 0 & 3 & 173 & .578 \\
    peri & 1 & 1 & 0 & 0 & 0 & 1 & {\bf 0} & 0 & 0 & 0 & 3 & .000 \\
    pra & 1 & 1 & 0 & 0 & 1 & 0 & 0 & {\bf 1} & 0 & 0 & 4 & .250 \\
    se & 7 & 6 & 0 & 0 & 2 & 2 & 0 & 0 & {\bf 8} & 2 & 27 & .296 \\
    teR & 15 & 3 & 5 & 0 & 1 & 3 & 0 & 0 & 2 & {\bf 214} & 243 & .881 \\
    \bottomrule
    \end{tabular}
}
    \begin{tablenotes}
        \small
        \item {\it Note.}The prefix in each row represents the prefix of a word and the prefix in each column represents the predicted prefix of a word. The bolded values represent the number of correct classifications of the LDA model. The overall accuracy is .88
    \end{tablenotes}
\end{table}

\subsubsection{Correlation Measures}\label{subsubsec3}

Three measures of semantic transparency were further calculated for each word. These are the correlation between each word and its prefix centroid (i.e., the mean vector of all words containing a particular prefix), between the vector of a word and its predicted vector derived from the Functional Representations of Affixes in Compositional Semantic Space (FRACSS; \citealp{marelli2015affixation}) model, and between the derived and shift vectors for each word.

\setlength{\parskip}{16pt}

In distributional semantics, the more two words occur in similar contexts, the smaller the cosine of the angle between their vectors in semantic space, and the greater their overlap in meaning. Alternatively, a very similar measure that tends to be highly correlated with cosine similarity is the Pearson correlation between two vectors. A greater cosine similarity corresponds to a stronger correlation between two vectors. As such, the correlation between each derived word and its centroid estimates the similarity in meaning between each word and all other words containing the same prefix.

\setlength{\parskip}{16pt}

The correlation estimates derived from the FRACSS model (\citealp{marelli2015affixation}) represent the similarity in meaning of the derived word and the predicted word. The FRACSS model proposed a linear mapping between the vectors of a derived word (e.g., revisit) and those of its base (e.g., visit). To implement FRACSS, the first step is to calculate the linear transformation that maps the vectors of a base word onto their corresponding derived words. This can be done by multiplying the matrix of the word vectors and the inverse of that of their base words. The next step is to calculate the predicted vectors of each word by multiplying the linear transformation with the vectors of the corresponding base words. For a step-by-step code on the implementation of the FRACSS model and a detailed discussion of FRACSS, see a JudiLing tutorial by \citet{heitmeier2024discriminative}. To compute the correlation estimates derived from FRACSS, the predicted vectors of each word are correlated with the vectors of the same word extracted from FastText. In simpler terms, the FRACSS correlation estimates represent the accuracy of the model in predicting a word, such that larger correlation coefficients indicate a more precise mapping for a derived word that is informed by its base.

\setlength{\parskip}{16pt}

Recall that the shift vector represents the displacement of a derived word from its base. To illustrate, on the top panel of Figure 2, a larger angle between the derived word and the shift vectors of a particular word corresponds to a smaller angle between the derived and its base vectors, thereby suggesting a smaller displacement of the derived form from its base as they appear closer to each other in semantic space. In contrast, on the bottom panel of Figure 2, a smaller angle (or stronger correlation) between the derived and shift vectors of a particular word corresponds to a larger angle (or weaker correlation) between the derived and its base vectors. In such a case, there is greater displacement of the derived word from its base, and are thus, semantically dissimilar as they appear far apart in semantic space. In our dataset, the two pairs of vectors (i.e., derived-shift, and derived-base) are strongly correlated, \textit{r} = -.7. 

\begin{figure}[H] 
    \centering
    \label{fig2}
    \includegraphics[width=0.8\textwidth]{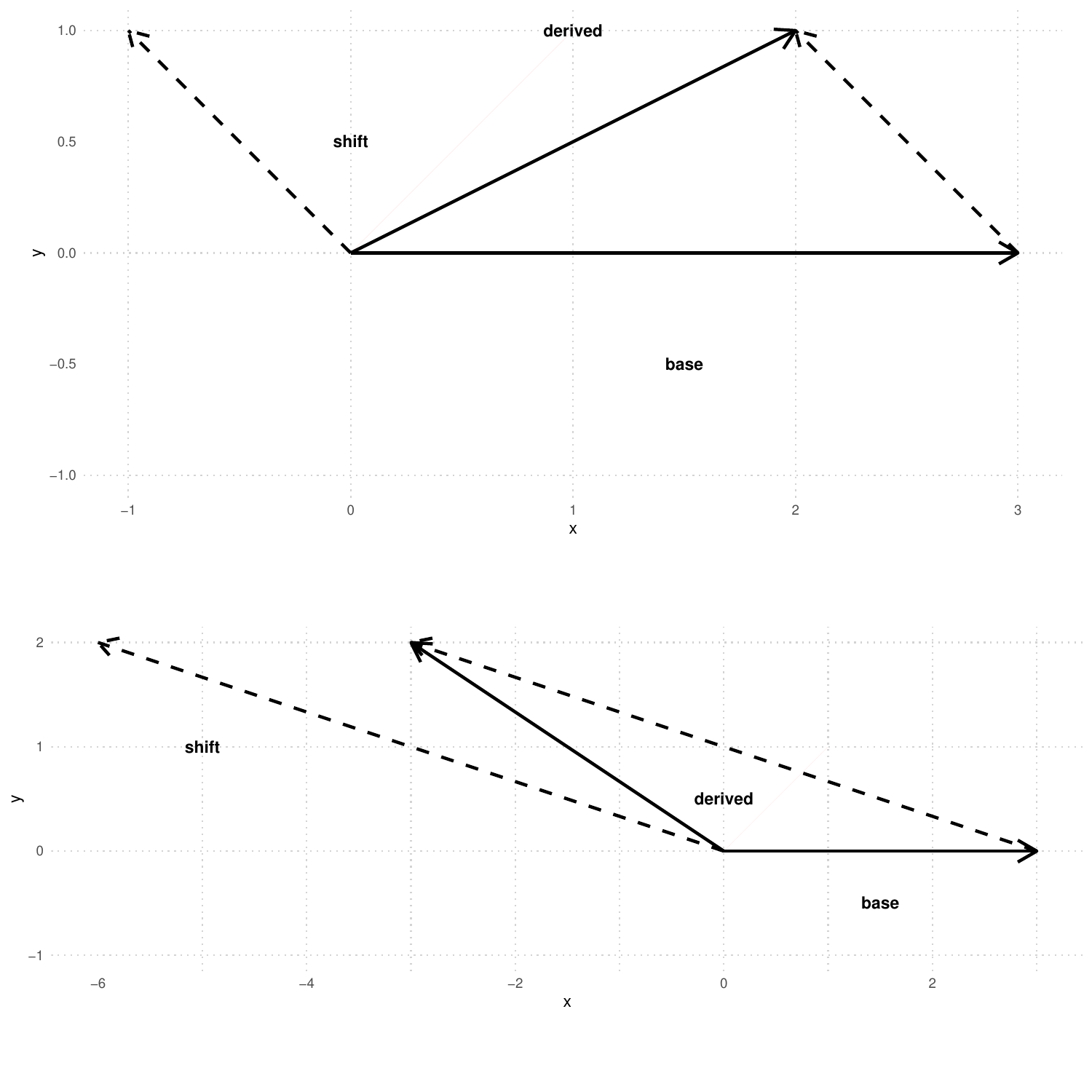}
    \caption{{\it Note.} Vector illustration. Solid lines indicate vectors for the base and derived word, and the dotted line represent the shift vectors. Of interest is the angle between the base and derived vectors, and the derived and shift vectors from the point of origin at (0, 0).}
\end{figure}

\subsection{Evaluation of Measures on Behavioural Data}\label{subsec3}

Each measure of interest was entered as a predictor, one at a time, in a series of Generalized Additive Mixed Models (GAMM) in R \citep{R} using the \textit{bam} function from the \textit{mgcv} package (\citealp{Wood:2017}) with the directive discrete set to TRUE. Both \textit{bam} and \textit{discrete=TRUE} make fitting a GAM model to a dataset much faster. Lexical decision latencies for 1,719 words from 280 participants were extracted from previous experiments of Malay visual word recognition (\citealp{maziyah2023malay}; \citealp{maziyah2023distributional}; \citealp{maziyah2025malay}). Only correct responses and RTs between 350ms and 3000ms that were within 2.5 SDs from the overall mean RTs were analyzed, yielding a total of 42,934 observations. Below we report whether each proposed measure was a significant predictor of RT and compared the effectiveness of each measure as predictors of Malay word recognition, with careful considerations of model residuals and concurvity statistics for model interpretability (see Supplementary Materials).

\setlength{\parskip}{16pt}

For comparison, we first ran a baseline model without the predictors of interest (see Table 3 and Figure 3). Whole-word frequency, word length, morphological root family size and its interaction with word frequency were entered as predictors. These predictors are shown to be crucial for Malay word recognition in previous studies of the Malay Lexicon Project (\citealp{maziyah2023malay}; \citealp{maziyah2023distributional}; \citealp{maziyah2025malay}). Word frequency and root family size were log-transformed, and a \textit{te} tensor product smooth was used to account for the main effect each of frequency and root family size, and their interaction. Random effects included trial number (centered and scaled) and subjects, using a factor smooth interaction \textit{bs} = “\textit{fs}” and a shrinkage directive \textit{m=1}. This structure of random effects, adopted from \citet{chuang2021analyzing} and \citet{baayen2022note}, is analogous to a random effects structure in a linear mixed model that has by-subject random intercepts and random slopes for trials. Results revealed a significant interaction between whole-word frequency and root family size, such that a facilitative effect of root family size on RT was most evident for lower frequency words. A facilitative effect of word frequency was observed across a large range of root family sizes. In addition, an inhibitory effect of word length was observed, particularly for words that were very long (\texttt{>}11 letters). These effects of whole-word frequency, root family size, and word length on RT were consistently observed in subsequent models.

\begin{table}[h]
    \centering
    \caption{GAMM - Baseline}\label{tab3}
    \begin{threeparttable}
        \resizebox{\textwidth}{!}{
            \begin{tabular*}{\textwidth}{@{\extracolsep{\fill}} l r r r r }
            \toprule
            {\bf Parametric coefficients} \\
            \hline
            Variable & Estimate  & Std. Error & {\it t}  &  {\it p} \\
            \hline
            Intercept & -1.18   & .03  & -44.90 & \texttt{<}.0001\\
            ExpNo2    & -0.21   & .03  & -7.18  & \texttt{<}.0001\\
            ExpNo3    & -0.14   & .03  & -4.62  & \texttt{<}.0001\\
            \hline
            {\bf Smooth terms} \\
            \hline
            Variable & edf  & Ref.df & F  &  {\it p} \\
            \hline
            Frequency*Family Size  & 13.84  & 16.57  & 146.01 & \texttt{<}.0001\\
            Word Length            & 6.47   & 7.49   & 154.52 & \texttt{<}.0001\\
            TrialNo, Subjects      & 863.38 & 2518.00 & 7.95   & \texttt{<}.0001\\
            \hline
            $R^2$ = .417 \\
            AIC = 12415.81 \\
            \hline
            \end{tabular*}
        }
        \begin{tablenotes}
            \small
            \item {\it Note.} Word frequency and root family size were log transformed. The model syntax is inverse RT $\sim~$ te(frequency*family size) + s(word length) + experiment + s(trial number, subjects, bs= ‘fs’, m=1). Inverse RT = -1000/RT; a negative sign is used to make the interpretability more like traditional RT data. 
        \end{tablenotes}
    \end{threeparttable}
\end{table}

\begin{figure}[H] 
    \centering
    \includegraphics[width=1.0\textwidth]{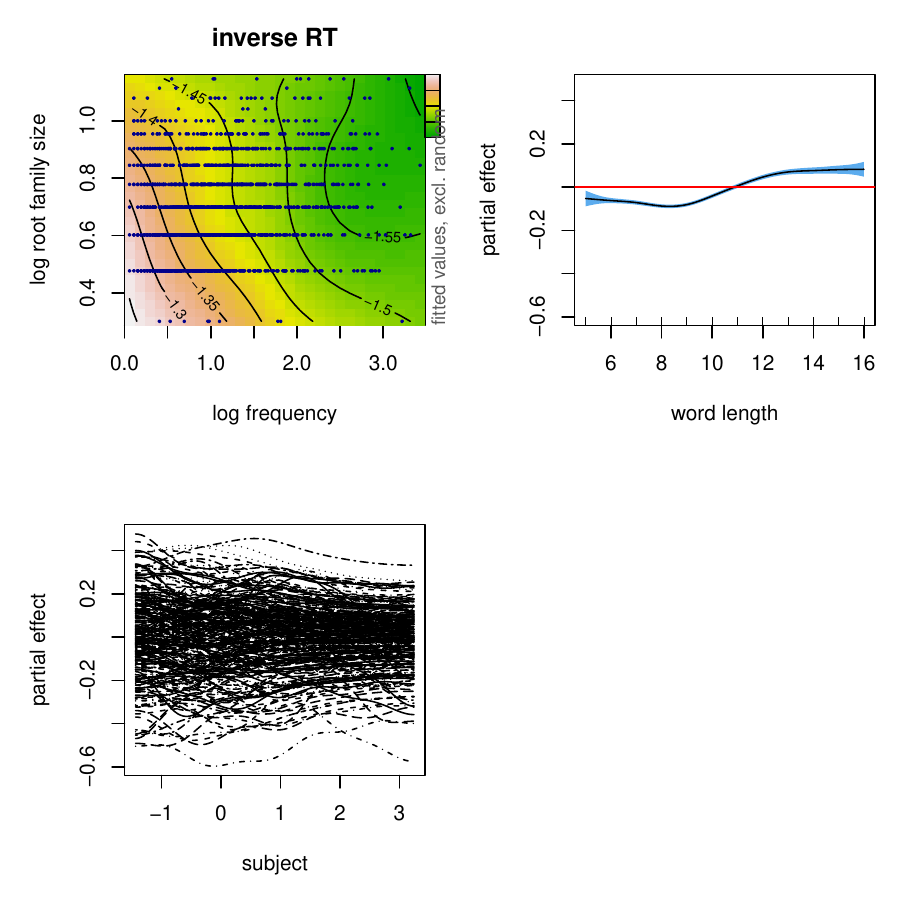}
    \caption{{\it Note.} Top row: Interaction between frequency and root family size (left). Data represented by dark blue points. Warmer colours (e.g., pink, orange) on the left-hand side denote longer RTs and cooler colours (e.g., green) on the right-hand side denote shorter RTs. Numbers on contour lines represent fitted inverse RT values. Partial effect of word length (right), with rugged lines on the x-axis representing the distribution of the data. Bottom row: Partial effects of trial number (centered and scaled) by subjects (left)}
    \label{fig3}
\end{figure}

An additional five GAM models were run, each with all terms in the baseline model and one measure of interest at a time. The two LDA classification scores were each entered as a linear predictor because there were only 10 unique values, one for each prefix. All three correlation estimates were entered as predictors using a thin plate regression spline smooth. All five measures were significant predictors of RT. Most crucially, each model that included the predictors of interest provided a better fit to the data than the baseline model (AIC = 12415.81).

\setlength{\parskip}{16pt}

A facilitative effect of the LDA accuracy was observed on RT. The greater the proportion of correct classifications for a particular prefix, the faster the responses. The accuracy of the LDA model in which shift vectors were used as input was a better predictor of RT (AIC = 12398.76; see Table 5) than the accuracy of the LDA that used the embeddings (AIC = 12406.96; see Table 4). 

\setlength{\parskip}{16pt}

The effect of the correlation between each derived word and their centroid (AIC = 12325.08; see Table 6) on RT was facilitative, except for a relatively small number of words that were very strongly correlated with their centroids (r \texttt{>}.7 and above; see left panel of Figure 4). For a large majority of the data, the closer a word is to all other words that share the same prefix, the more easily it is recognized, as observed by shorter RTs. If, however, a word is too close in semantic space to all other words that share the same prefix, then word recognition appears more effortful, as observed by longer RTs. It is possible that processing is more effortful because such a word appears more confusable with its morphologically related words. 

Additionally, we observed a facilitative effect of the FRACSS correlation estimates on RT (AIC = 12340.77; see Table 7). Higher estimates indicate a more precise prediction of the derived word. Faster responses were observed for words that were predicted more accurately by FRACSS (see middle panel of Figure 4). In contrast, we observed an inhibitory effect of the correlation between the derived and shift vectors of each word (AIC = 12381.06; see Table 8). As mentioned earlier, a stronger correlation between the derived and shift vectors of a word indicate a greater displacement of the derived form from its base. In such cases, slower responses are elicited (see right panel of Figure 4). Across all five measures, the model that included the correlation between the vectors of the derived word and its corresponding centroid provided the best fit to the data (see Table 9). 

\clearpage

\begin{table}[h]
    \centering
    \caption{GAMM - LDA (Embeddings)}\label{tab4}
    \begin{threeparttable}
        \resizebox{\textwidth}{!}{
            \begin{tabular*}{\textwidth}{@{\extracolsep\fill}lrrrrrr}
            \toprule
            {\bf Parametric coefficients}\\
            \hline
            Variable & Estimate  & Std. Error & {\it t}  &  {\it p}\\
            \hline
            Intercept     & -1.14   & .03  & -38.02 & \texttt{<}.0001\\
            LDA-Embedding & -0.05   & .02  & -3.40  & .0007\\
            ExpNo2        & -0.21   & .03  & -7.18  & \texttt{<}.0001\\
            ExpNo3        & -0.14   & .03  & -4.58  & \texttt{<}.0001\\
            \hline
            {\bf Smooth terms}\\
            \hline
            Variable & edf  & Ref.df & F  &  {\it p}\\
            \hline
            Frequency*Family Size  & 13.66  & 16.38  & 148.40 & \texttt{<}.0001\\
            Word Length            & 6.45  & 7.47    & 156.46 & \texttt{<}.0001\\
            TrialNo, Subjects      & 863.22 & 2518.00& 7.95   & \texttt{<}.0001\\
            \hline
            $R^2$ = .418\\
            AIC = 12406.96\\
            \hline
            \end{tabular*}
        }
        \begin{tablenotes}
            \small
            \item {\it Note.} Word frequency and root family size were log transformed. The model syntax is inverse RT $\sim~$ te(frequency*family size) + s(word length) + LDA-Embedding + experiment + s(trial number, subjects, bs= ‘fs’, m=1). Inverse RT = -1000/RT; a negative sign is used to make the interpretability more like traditional RT data
        \end{tablenotes}
    \end{threeparttable}
\end{table}

\begin{table}[h]
    \centering
    \caption{GAMM - LDA (Shift)}\label{tab5}
    \begin{threeparttable}
        \resizebox{\textwidth}{!}{
            \begin{tabular*}{\textwidth}{@{\extracolsep\fill}lrrrrrr}
            \toprule
            {\bf Parametric coefficients}\\
            \hline
            Variable & Estimate  & Std. Error & {\it t}  &  {\it p}\\
            \hline
            Intercept & -1.14   & .03  & -40.66 & \texttt{<}.0001\\
            LDA-Shift & -0.05   & .01  & -4.49 & \texttt{<}.0001\\
            ExpNo2    & -0.21   & .03  & -7.17  & \texttt{<}.0001\\
            ExpNo3    & -0.14   & .03  & -4.55  & \texttt{<}.0001\\
            \hline
            {\bf Smooth terms}\\
            \hline
            Variable & edf  & Ref.df & F  &  {\it p}\\
            \hline
            Frequency*Family Size  & 13.52  & 16.24  & 150.19 & \texttt{<}.0001\\
            Word Length            & 6.47   & 7.49   & 157.36 & \texttt{<}.0001\\
            TrialNo, Subjects      & 863.21 & 2518.00& 7.95   & \texttt{<}.0001\\
            \hline
            $R^2$ = .418\\
            AIC = 12398.76\\
            \hline
            \end{tabular*}
        }
        \begin{tablenotes}
            \small
            \item {\it Note.} Word frequency and root family size were log transformed. The model syntax is inverse RT $\sim~$ te(frequency*family size) + s(word length) + LDA-Shift + experiment + s(trial number, subjects, bs= ‘fs’, m=1). Inverse RT = -1000/RT; a negative sign is used to make the interpretability more like traditional RT data
        \end{tablenotes}
    \end{threeparttable}
\end{table}

\begin{table}[h]
    \centering
    \caption{GAMM - Correlation (Derived Word and Centroid)}\label{tab6}
    \begin{threeparttable}
        \resizebox{\textwidth}{!}{
            \begin{tabular*}{\textwidth}{@{\extracolsep\fill}lrrrrrr}
            \toprule
            {\bf Parametric coefficients}\\
            \hline
            Variable & Estimate  & Std. Error & {\it t}  &  {\it p}\\
            \hline
            Intercept & -1.18   & .03  & -44.79 & \texttt{<}.0001\\
            ExpNo2    & -0.21   & .03  & -7.20  & \texttt{<}.0001\\
            ExpNo3    & -0.14   & .03  & -4.61  & \texttt{<}.0001\\
            \hline
            {\bf Smooth terms}\\
            \hline
            Variable & edf  & Ref.df & F  &  {\it p}\\
            \hline
            Frequency*Family Size     & 13.41  & 16.06  & 146.12 & \texttt{<}.0001\\
            Word Length               & 6.47   & 7.49   & 158.03 & \texttt{<}.0001\\
            Correlation-dev.centroid  & 6.13   &  7.30  & 13.24  & \texttt{<}.0001\\
            TrialNo, Subjects         & 864.68 & 2518.00& 7.97   & \texttt{<}.0001\\
            \hline
            $R^2$ = .419\\
            AIC = 12325.08\\
            \hline
            \end{tabular*}
        }
        \begin{tablenotes}
            \small
            \item {\it Note.} Word frequency and root family size were log transformed. The model syntax is inverse RT $\sim~$ te(frequency*family size) + word length + s(correlation-dev.centroid) + experiment + s(trial number, subjects, bs= ‘fs’, m=1). Inverse RT = -1000/RT; a negative sign is used to make the interpretability more like traditional RT data
        \end{tablenotes}
    \end{threeparttable}
\end{table}

\begin{table}[h]
    \centering
    \caption{GAMM - Correlation (Target Word and Predicted Word; FRACSS)}\label{tab7}
    \begin{threeparttable}
        \resizebox{\textwidth}{!}{
            \begin{tabular*}{\textwidth}{@{\extracolsep\fill}lrrrrrr}
            \toprule
            {\bf Parametric coefficients}\\
            \hline
            Variable & Estimate  & Std. Error & {\it t}  &  {\it p}\\
            \hline
            Intercept & -1.18   & .03  & -45.02 & \texttt{<}.0001\\
            ExpNo2    & -0.22   & .03  & -7.29  & \texttt{<}.0001\\
            ExpNo3    & -0.14   & .03  & -4.69  & \texttt{<}.0001\\
            \hline
            {\bf Smooth terms}\\
            \hline
            Variable & edf  & Ref.df & F  &  {\it p}\\
            \hline
            Frequency*Family Size     & 14.00  & 16.73  & 147.08 & \texttt{<}.0001\\
            Word Length               & 6.38   & 7.40   & 129.71 & \texttt{<}.0001\\
            Correlation-FRACSS        & 5.52   &  6.63  & 12.35  & \texttt{<}.0001\\
            TrialNo, Subjects         & 862.79 & 2518.00& 7.97   & \texttt{<}.0001\\
            \hline
            $R^2$ = .419\\
            AIC = 12340.77\\
            \hline
            \end{tabular*}
        }
        \begin{tablenotes}
            \small
            \item {\it Note.} Word frequency and root family size were log transformed. The model syntax is inverse RT $\sim~$ te(frequency*family size) + word length + s(correlation-FRACSS) + experiment + s(trial number, subjects, bs= ‘fs’, m=1). Inverse RT = -1000/RT; a negative sign is used to make the interpretability more like traditional RT data
        \end{tablenotes}
    \end{threeparttable}
\end{table}

\begin{table}[h]
    \centering
    \caption{GAMM - Correlation (Derived Word and Shift)}\label{tab8}
    \begin{threeparttable}
        \resizebox{\textwidth}{!}{
            \begin{tabular*}{\textwidth}{@{\extracolsep\fill}lrrrrrr}
            \toprule
            {\bf Parametric coefficients}\\
            \hline
            Variable & Estimate  & Std. Error & {\it t}  &  {\it p}\\
            \hline
            Intercept & -1.19   & .03  & -45.18 & \texttt{<}.0001\\
            ExpNo2    & -0.21   & .03  & -7.08  & \texttt{<}.0001\\
            ExpNo3    & -0.14   & .03  & -4.48  & \texttt{<}.0001\\
            \hline
            {\bf Smooth terms}\\
            \hline
            Variable & edf  & Ref.df & F  &  {\it p}\\
            \hline
            Frequency*Family Size     & 13.57  & 16.73  & 134.90 & \texttt{<}.0001\\
            Word Length               & 6.49   & 7.40  & 153.90  & \texttt{<}.0001\\
            Correlation-dev.shift     & 2.88   &  6.63  & 10.47  & \texttt{<}.0001\\
            TrialNo, Subjects         & 863.84 & 2518.00& 7.96   & \texttt{<}.0001\\
            \hline
            $R^2$ = .419\\
            AIC = 12381.06\\
            \hline
            \end{tabular*}
        }
        \begin{tablenotes}
            \small
            \item {\it Note.} Word frequency and root family size were log transformed. The model syntax is inverse RT $\sim~$ te(frequency*family size) + word length + s(correlation-dev.shift) + experiment + s(trial number, subjects, bs= ‘fs’, m=1). Inverse RT = -1000/RT; a negative sign is used to make the interpretability more like traditional RT data
        \end{tablenotes}
    \end{threeparttable}
\end{table}

\clearpage 

\begin{figure}[H] 
    \centering
    \includegraphics[width=1.0\textwidth]{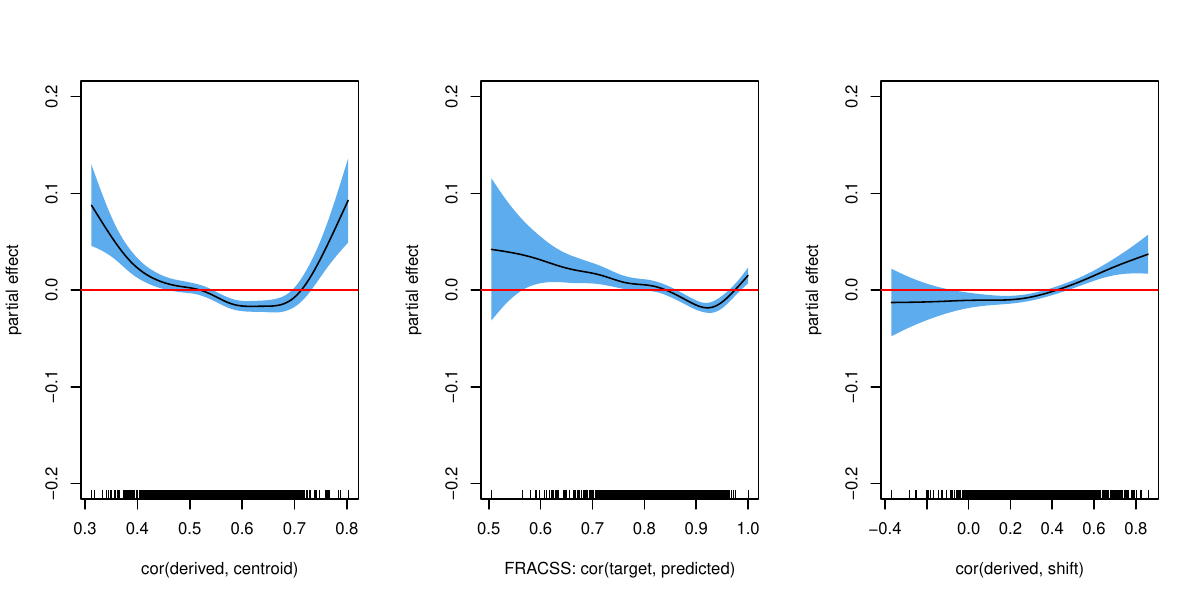}
    \caption{{\it Note.} Partial effect of each correlation measure on RT. Black lines on the x-axis of each plot represent the data}
    \label{fig4}
    \end{figure}

\begin{table}[H]
    \centering
    \caption{AIC Comparison}\label{tab9}
    \begin{threeparttable}
        \resizebox{\textwidth}{!}{
            \begin{tabular*}{\textwidth}{@{\extracolsep\fill}lrr}
            \toprule
            {\bf Model} & $\Delta$ AIC\\
            \hline
            LDA-Derived & 8.85 \\
            LDA-Shift    & 17.05 \\
            Correlation-Dev.centroid   & 90.73  \\
            Correlation-FRACSS   & 75.04  \\
            Correlation-Shift.centroid   & 34.75  \\
            \hline
            \end{tabular*}
            }
        \begin{tablenotes}
            \small
            \item {\it Note.} Difference in AIC scores are calculated by subtracting the AIC of each model that included a predictor of interest from the AIC of the baseline model. Greater values in the change of AIC score indicate a better fit to the data.
        \end{tablenotes}
    \end{threeparttable}
\end{table}

\subsection{General Discussion}\label{subsec3}

\setlength{\parskip}{16pt}

In the present study, we first sought to explore whether the embeddings of Malay complex words cluster in semantic space by prefix, using t-SNE. To make any meaningful interpretation, we focus on words containing a prefix of a sizeable count. Namely, these are words that contained the prefix \textit{beR-, meN-, teR-, peN-, peR-,} or \textit{ke-}. A key take-away from the t-SNE analysis is the observation that there is considerable variation between complex words that arise from prefixes in Malay, a stark contrast to polysemous particles in German that are semantically ambiguous \citep{stupak2022inquiry}. More broadly, we have demonstrated that inquiries into the semantic transparency of words can be meaningfully explored using \textit{t}-SNE.

A detailed linguistic analysis of how these clusters emerge as depicted in the \textit{t}-SNE plot is beyond the scope of the present study. As an aside, however, we inspected whether clustering occurs by word category. Across languages, many previous studies have suggested that nouns and verbs differ in their semantics and distributional properties (for a review, see \citealp{vigliocco2011nouns}). We extracted word category information from the msTenTen corpus, a Malay web corpus, on SketchEngine for each word in our dataset. A large majority of the words in our dataset were assigned as nouns. Some clustering was observed for verbs, although they appear largely nested in the cluster of nouns (see Figure S1 in Supplementary Materials). No obvious clusters were observed for adjectives. Relatedly, the LDA model correctly classified nouns and verbs to a large extent, but not for adjectives. The overall accuracy with which the LDA model predicts word category is much lower (.77) than the LDA models that predict a word's prefix (.93 using word embeddings and .88 using shift vectors). Most crucially, the clustering reported for prefixed words in the present study is not confounded with word category. 

A next step would be to identify a set of features in which these clusters embody. Such a study will shed light on the kind of semantics that could be extracted from word embeddings. We leave more fine-grained analyses for future work. For an initial exploratory study on prefixed words, it makes most sense to first examine whether or not complex words cluster meaningfully by their prefixes. The results of the present study make it sufficiently clear that word embeddings capture a rich knowledge of semantic information that could be used to discriminate between complex words.

\setlength{\parskip}{16pt}

A secondary objective of the present study was to calculate several measures of semantic transparency and evaluated their impact on lexical decision latencies in a series of GAM models. All five measures significantly predicted decision latencies above and beyond classical predictors of lexical processing and root family size. This finding complements prior work in German \citep{stupak2022inquiry} and Mandarin \citep{shen2022adjective} in which distinct clusters in semantic space were formed for words that share a derivational affix and show a strong semantic association with their base words. In the present study, the correlation between a word and its corresponding centroid emerged as the best predictor of decision latencies.

Unlike word embeddings, the vector of a centroid does not represent a real word, but rather, an average embedding that could be understood as the prototypical meaning of the prefix. In our case, words are considered related if they share a prefix. We showed that the speed with which a word is processed is predictable in part from the strength of the semantic relationship between a particular word and all other related words, such that faster responses were elicited for a word that shares a strong semantic relationship with its centroid. Such evidence provide empirical support for the potential of realizational morphology as a theory of lexical processing, even for Malay, a language that is morphologically rich in derivation and contains minimal inflection. Although subword embeddings are considered, morphemes are not explicitly represented. Realizational morphology is typically discussed in studies concerning inflected forms. Only derived forms were analyzed in this study. These results further support a previous exploratory study on Indonesian morphology \citep{denistia2022morphology} that used the Discriminative Lexicon Model (DLM; \citealp{baayen2019discriminative}. The DLM, grounded in word and paradigm morphology, is a computational theory of the mental lexicon in which the whole word is taken to be the most basic unit. The DLM consists of simple linear mappings between high-dimensional representations of form and meaning (for details on the implementation of the DLM, see JudiLing tutorial; \citealp{heitmeier2024discriminative}). In that study of Indonesian morphology, the DLM accurately discriminated between words containing prefixes \textit{pe-} and \textit{pen-}, often associated with similar meanings, even though the model was not explicitly informed about exponents and stems. On our to-do list is a closer inspection of the potential role of centroids in the DLM. Preliminary evidence from the DLM trained on words in the present study's dataset revealed that there is indeed a strong correspondence between the centroid and the linear mappings of a word's form and its meaning (see Figure 5), further justifying a promising role of centroids in word recognition.

\begin{figure}[H] 
    \centering
    \label{fig5}
    \includegraphics[width=0.8\textwidth]{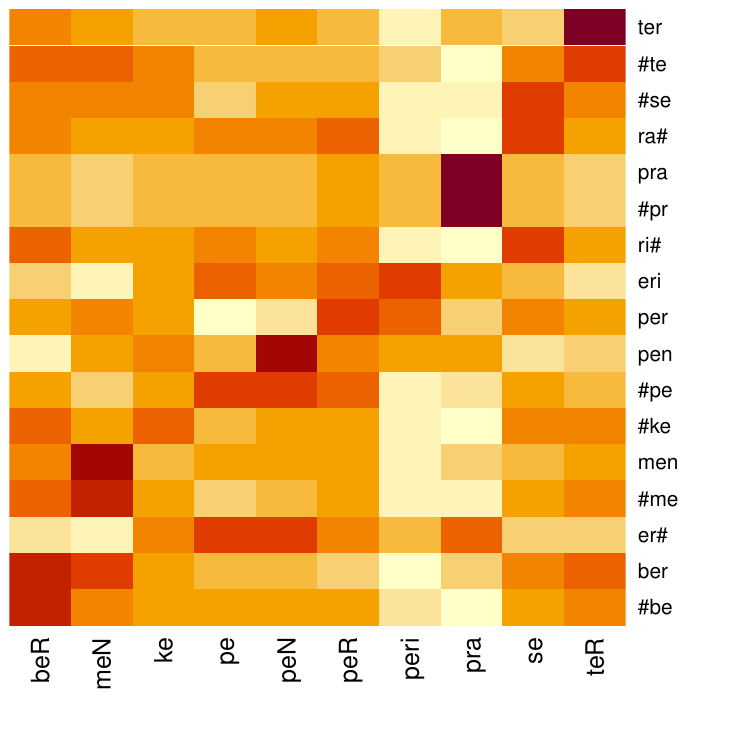}
    \caption{{\it Note.} Correlation heatmap of prefix centroid embeddings and a subset of the linear comprehension mapping F that maps the embeddings of a word form (trigrams) to its meaning (FastText) in the DLM. The set of trigrams presented correspond to at least one of the prefixes. Darker shades, compared to lighter shades, indicate a stronger correlation between the embeddings of the centroid and comprehension mapping F. The strongest correlations are present for the trigrams that correspond to the prefix, indicating that it is the prefixal trigrams that contribute most to realizing the meaning of the centroid}
\end{figure}

Furthermore, each of the three models that included a correlation measure provided a better fit to the data than either of the two models that included a measure derived from the LDA, providing support for analogical patterns in contrast to strict classification. These findings resonate with the results of several studies by Westbury and colleagues. In \citet{westbury2023human}, both human animacy judgments and word embedding models are shown to produce good approximations of animacy ratings on the basis of family resemblance rather than distinct category memberships. For instance, words related to human beings such as \textit{professors} received an animacy rating of .60 by human participants and .55 by word embedding models, providing more support for the idea of similarity compared to binary classifications even for a concept as basic as animacy. \citet{westbury2019conceptualizing} and \cite{westbury2019wriggly} successfully demonstrated that the centroid, in their case, the mean vector of words in a particular word category can be used to assess category membership such as nouns, verbs, and adjectives. Words of a particular category are highly correlated with their centroid. The authors noted that their findings regarding centroids extend to many semantic properties. In the present study, we establish that the centroid can be used to assess semantic transparency of a prefix in Malay.

\setlength{\parskip}{16pt}

Between the two LDA-based measures, the model that included the classification scores derived from the shift vectors provided a better fit to the data than the model that included the classification scores derived from just the word embeddings. The shift vectors represent a snapshot of the transformation in meaning of the derived form from its base. Recent work on English has shown that the semantics of pluralization varies by semantic class, even though such differences are not marked morphologically \citep{shafaei2024pluralization}. For instance, the nature of the change in semantic space from singular to plural differs between words that describe a person or an animal. In that study, using shift vectors, distinct clusters in a \textit{t}-SNE were observed for a large set of WordNet supersenses that include broad semantic categories for nouns. Our findings lend support to prior work in that such movements through semantic space, up to the point that the meaning of a derived form is realized, account for additional information that is meaningful for word recognition.

\section{Conclusion}\label{sec4}
The present study reports an exploratory analysis of the semantic geometry of Malay word embeddings in high dimensional space. Techniques used in machine learning were employed in the visualization of word embeddings for ease of interpretability and in the computation of embedding-based measures of semantic transparency. We observed distinct clusters of complex words varied by their prefix class. In addition, we provide evidence that each embedding-based measure significantly predicts lexical decision latencies. In particular, the model that included the correlation between each derived word and their centroid appears to be the best fit to the data. That is, the similarity of each word to the prototypical meaning of its prefix appears to be the best way to characterize semantic transparency. These measures are available for a large set of Malay words and can be downloaded in the latest version of the Malay Lexicon Project 3: \url{https://osf.io/dhyzb/?view_only=e05e71b31cb54daf94a55f46f9cc82da}

\bibliography{data}
\end{document}